%% file: main.tex
\documentclass[conference]{ieeeconf}
\usepackage{graphicx}
\usepackage{booktabs,tabularx}
\usepackage{changepage}
\usepackage{amsmath}
\usepackage{threeparttable}
\usepackage{booktabs}
\usepackage{makecell}
\usepackage[dvipsnames,svgnames,x11names]{xcolor}
\usepackage[ruled,vlined,noend]{algorithm2e}
\usepackage{afterpage}

\usepackage{stfloats}
\usepackage{float}
\usepackage{cite}
\usepackage{amssymb}
\usepackage{comment}
\usepackage[acronym]{glossaries}
\usepackage{glossaries-extra}
\usepackage[dvipsnames,svgnames,x11names]{xcolor}
\usepackage[hidelinks]{hyperref}
\usepackage[normalem]{ulem}

\newcommand{\graydash}{{\color{lightgray}--}}

\usepackage[normalem]{ulem}
\usepackage{tikz}
\usepackage{xcolor}

\newcommand{\colorunderline}[2]{%
  \tikz[baseline=(text.base)]{
    \node[inner sep=0pt, outer sep=0pt] (text) {#2};
    \draw[#1, line width=0.6pt] ([yshift=-0.8pt]text.south west) -- ([yshift=-0.8pt]text.south east);
  }%
}

\newcommand{\colordashunderline}[2]{%
  \tikz[baseline=(text.base)]{
    \node[inner sep=0pt, outer sep=0pt] (text) {#2};
    \draw[#1, dashed, dash pattern=on 2pt off 1.5pt, line width=0.6pt] ([yshift=-0.8pt]text.south west) -- ([yshift=-0.8pt]text.south east);
  }%
}

\definecolor{myopicColor}{HTML}{e65f25}
\definecolor{selfishColor}{HTML}{3f76bb}
\definecolor{courteousColor}{HTML}{51be9a}
\definecolor{courteousJointColor}{HTML}{51be9a}

\usepackage[table]{xcolor}
\definecolor{colMyopic}{HTML}{faded1}
\definecolor{colSelfAP}{HTML}{b2c9e5}
\definecolor{colJointAP}{HTML}{ccefe4}
\definecolor{colOursAP}{HTML}{ccefe4}
\definecolor{colDef}{HTML}{ffffff}

\setabbreviationstyle[acronym]{long-short}
\setabbreviationstyle[short]{short-nolong}
\newacronym[category={short}]{pddl}{\textsc{pddl}}{The Planning Domain Definition Language}
\newacronym{gnn}{\textsc{gnn}}{graph neural network}
\newacronym[category={short}]{3dsg}{3D scene graph}{3D Scene Graph}
\newacronym[category={short}]{procthor}{\textsc{ProcTHOR}}{ProcTHOR}
\newacronym[category={short}]{alfred}{\textsc{alfred}}{Action Learning From Realistic Environments and Directives}

\newcommand{\apcost}{V_\textsc{A.P.}}

\IEEEoverridecommandlockouts

\title{\LARGE \bf Courteous Anticipation: Improving Long-Lived Task Planning in Persistent Shared Environments}

\author{Md Ridwan Hossain Talukder, Elizabeth Phillips and Gregory J. Stein%
\thanks{Md Ridwan Hossain Talukder and Gregory J. Stein are affiliated with the CS Department and Elizabeth Phillips is affiliated with the Department of Psychology, George Mason University, Fairfax, VA: {\tt\small \{mtalukd, ephill3, gjstein\}@gmu.edu}}}%

\begin{document}
\maketitle
\input{Section/Abstract}

\input{Section/Introduction}
\input{Section/RelatedWork}
\input{Section/Problem}
\input{Section/Approach}
\input{Section/Methodology}
\input{Section/Experiment}
\input{Section/Conclusion}

\bibliographystyle{IEEEtran}
\bibliography{references.bib}

\end{document}

%% file: Section/Abstract.tex
\begin{abstract}
We consider a task planning scenario in which robots sharing a persistent environment are assigned tasks one at a time from a held-out sequence. 
Standard task planners, lacking foresight of future tasks and inconsiderate of others' constraints, solve each task in isolation, leaving terminal states that increase future cost for all—side effects that compound over lengthy task sequences.
To reduce cost over the sequence, a robot must anticipate how its actions now may impact performance on future tasks for all robots sharing the environment.
Therefore, we present \emph{courteous anticipatory planning}, wherein a model-based planner proposes candidate plans and selects the one that jointly minimizes immediate cost and aggregated expected future cost across all robots, estimated via independent per-robot learned estimators. 
This factored formulation avoids combinatorial joint rollouts and supports modular deployment: adding a robot requires only training its own estimator.
We evaluate in two persistent PDDL domains, a home environment with robots that have similar capabilities but different responsibilities, and a restaurant environment where robots' distinct capabilities create states that other robots lack the capability to resolve. 
During lengthy task sequences, our planner reduces total cost by 10.4\% versus myopic and 4.5\% versus selfish anticipatory planning in a two-robot home environment and by 20.5\% and 6.7\%, respectively, in a three-robot restaurant.

\end{abstract}

%% file: Section/Introduction.tex
\section{Introduction}

Robots are becoming an integral part of everyday life, assisting humans in daily chores and professional workspaces such as homes, restaurants, hospitals, and factories. 
As robots take on more responsibilities, they are increasingly expected to operate over long periods of time rather than being deployed for a single isolated task. 
In these shared environments, each robot's actions modify the environment in ways that influence the performance of all robots sharing the space.
What separates a good inhabitant of a shared space from a poor one is not what it can do, but whether it thinks about how its action impacts others.

\begin{figure}
    \includegraphics[width=8.5cm]{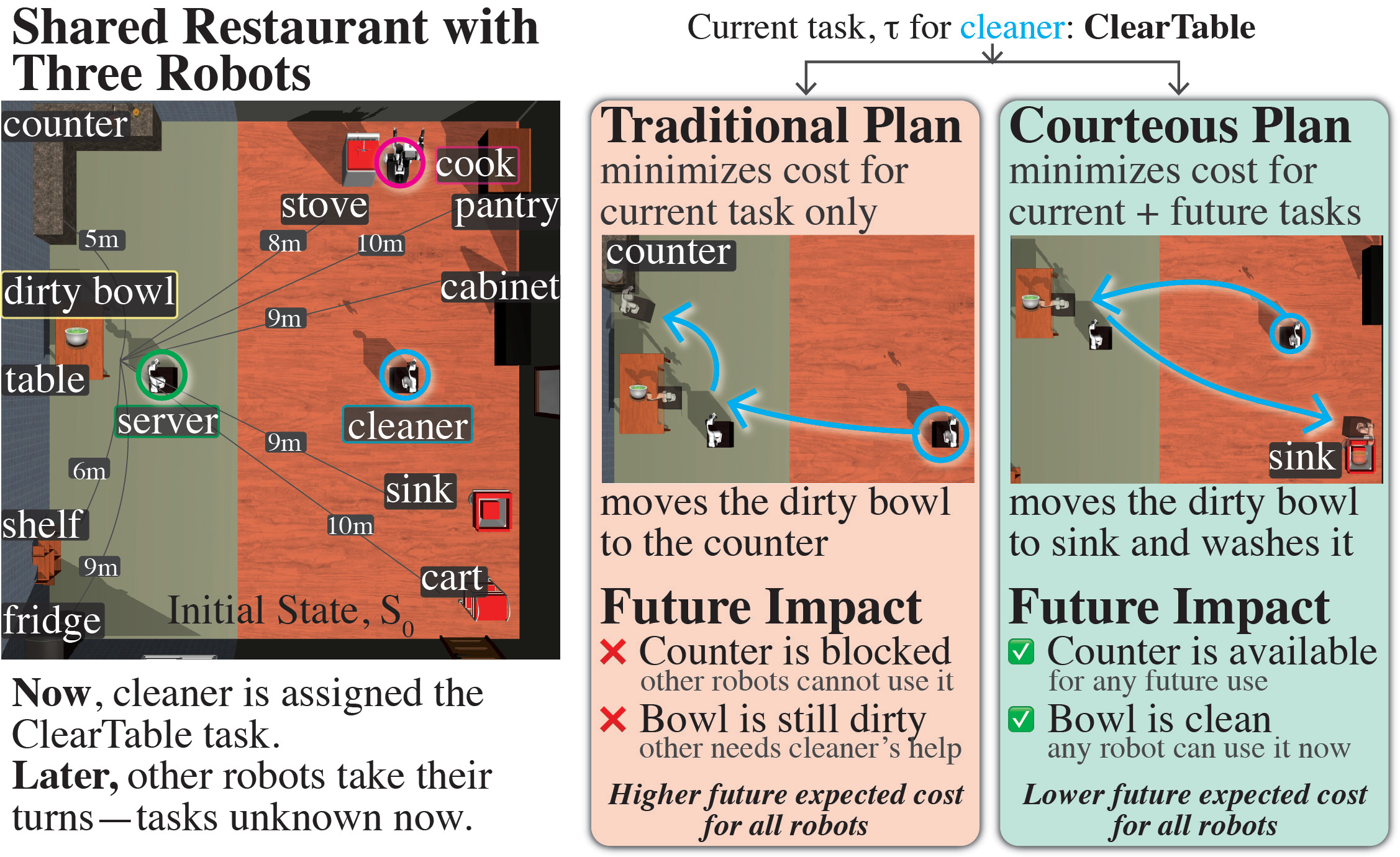}
    \centering
    \vspace{-0.5em}
    \caption{\textbf{Courteous Anticipation:} Given the task \texttt{ClearTable}, a \textbf{myopic} plan leaves the dirty bowl on the \textit{counter}, blocking all robots from using it and forcing others to request help from the cleaner whenever a later task requires a clean bowl. \textbf{Our Approach} anticipates the impact of this action on all robots and courteously cleans and places the dirty bowl near the \textit{sink}, reducing the overall planning cost.}
    \label{fig:example}
    \vspace{-1.5em}
\end{figure}

Consider a shared kitchen with three robots (Fig.~\ref{fig:example}): a tall cook that alone can safely access the stove, a short cleaner that alone can wash cookware and a short server that alone can serve meals and refill food items. 
Suppose the cleaner is assigned the \texttt{ClearTable} task which requires moving the dirty bowl from table.
Lacking foresight of the future tasks, a conventional plan~\cite{garrett2020pddlstream, plaku2010sampling,kaelbling2011hierarchical, srivastava2014combined, kim2020learning, kim2019learning,chitnis2016guided,dantam2016incremental, silver2021planning} for the cleaner focuses on completing the current task as quickly as possible.
This \emph{myopic} plan (Fig.~\ref{fig:example}, orange) clears the table by moving the dirty bowl to the nearest counter, inadvertently creating two costly side effects: it blocks any robot that requires a free counter; and it forces the cook and server to \emph{request help}, a costly action that activates an otherwise idle robot, from the cleaner before completing any task that requires a clean bowl.
Instead, the robot has an opportunity to be \emph{courteous} and reduce downstream friction; absent such courtesy myopic plan increases the overall cost of the sequence of tasks eventually assigned to each robot.

To exhibit courteous behavior in a shared environment, a robot must reason beyond its immediate task and anticipate how its actions \emph{now} will affect others as they are eventually assigned tasks of their own. 
In the earlier example, a courteous plan for the cleaner (Fig.~\ref{fig:example}, green) would clean the bowl and place it in the sink, freeing the counter for others and eliminating the need for later tasks to request the cleaner's help.
We call this behavior \emph{courteous anticipation}: the ability of a robot to complete its own task while being considerate of the future tasks and capabilities of others and acting accordingly. However, achieving courteous anticipation is challenging: a robot must act without knowing what tasks it or others will be assigned next and how it impacts others physical constraints. With the broad responsibilities of each robot, reasoning directly over all such tasks that may be assigned quickly becomes combinatorially expensive.

The emerging field of anticipatory planning \cite{dhakal2023anticipatory, talukder2025, task-anticipation} proposes an objective function in which a robot considers
both its immediate objective and the impact of its actions
on tasks it may be later assigned. However, in shared spaces with several robots, self-interested anticipation is insufficient. 
A robot that considers only for its own future performance may still leave the space at a terminal state that is costly for robots sharing the space: behavior we term as \textit{selfish} anticipation. 
To improve \emph{collective} behavior in shared spaces, especially when each robot has different responsibilities and distinct capabilities, a robot must consider the impact of its actions on others. It is a key insight of this work that standard task planning in shared spaces, absent this consideration, routinely creates negative side effects and increases the cost of later tasks of others. Our \emph{courteous anticipation} aims to mitigate these side effects by making the acting robot account for others, reducing the cost of sequential task planning.

We identify an important obstacle to effective sequential task planning in persistent shared environments: the pervasive absence of \emph{courtesy} from standard task planning in these domains introduces negative side effects that increase overall planning cost, an effect amplified when robots differ in capabilities and responsibilities.
To address this, we introduce \emph{courteous anticipatory planning}: a planning objective that accounts for the distinct capabilities and responsibilities of all robots by jointly minimizing the immediate cost of completing the current task and the expected future cost it induces for all robots sharing the space. 
Directly evaluating this expected future cost across all robots requires brute-force joint rollouts over the combinatorial space of possible task sequences, which becomes computationally intractable as robots and responsibilities grow. 
Instead, we decompose this expected future cost into independent per-robot estimates, each trained to capture how the environment state affect that robot's ability to fulfill its responsibilities under its structural constraints. At planning time, we aggregate these estimates to select the most courteous plan, avoiding joint rollouts and requiring no retraining of existing estimators when robots are added or removed. 

We evaluate our approach in shared-environment settings of increasing complexity: a home setting that establishes improvements from robots' distinct responsibilities alone, and a restaurant setting that demonstrates distinct capabilities amplify these gains, both evaluated against myopic and selfish anticipatory baselines. Evaluated over lengthy task sequences, our planner reduces total cost by 10.4\% versus myopic and 4.5\% versus selfish anticipatory planning in a two-robot home environment, and by 20.5\% and 6.7\%, respectively, in a three-robot restaurant.

%% file: Section/RelatedWork.tex
\section{Related Work} \label{sec:related-work}
\textbf{Task Planning and Learning-Augmented Planning}
Conventional task planning approaches~\cite{garrett2020pddlstream, plaku2010sampling, kaelbling2011hierarchical, srivastava2014combined} seek to solve each task at minimum immediate cost, without considering how the resulting terminal state affects future tasks. Some approaches augment planning with learned heuristics or value functions~\cite{kim2020learning, kim2019learning, chitnis2016guided, dantam2016incremental, silver2021planning} to accelerate the search, but not to reason about the downstream consequences of these solutions. In persistent shared environments, however, this myopia is costly: plans may leave a terminal state that increases task cost for robots with different structural constraints and responsibilities. Our approach instead selects among plans by reasoning about downstream consequences for all robots, using learned per-robot estimators that capture each robot's capabilities and responsibilities.

\textbf{Anticipating Future Tasks in Persistent Environments}
Recent work has presented approaches to anticipate and mitigate unforeseen side effects on future unseen tasks within a persistent physical space~\cite{dhakal2023anticipatory, talukder2025, task-anticipation}, estimating future costs using learned models to avoid expensive rollouts. Other techniques have also been developed to proactively anticipate the usage of objects to assist others~\cite{patel2023predicting, patel2022proactive}, to align human and robot expectations by designing environments~\cite{sikes2024reducing}, or to minimize divergence from a goal state~\cite{koppula2016anticipatory}. Online task planning scenarios have also been considered for anticipatory planning~\cite{burns2012anticipatory}, focusing on the temporal arrival of tasks for robots rather than sequential arrangement. Nevertheless, all these techniques are single-robot techniques; they only consider how a robot’s actions influence its future tasks but not how these actions limit or enable other robots with varying structural constraints and goals in a shared space. 
Klassen et al.~\cite{klassen2022planning} similarly motivate looking beyond an agent's own abilities, but their compilation-based approach does not extend to long-horizon task sequences: it minimizes side effects against a single, given future goal per agent, rather than a distribution over many future tasks across a persistent sequence.


%% file: Section/Problem.tex
\section{Problem Statement} \label{sec:problem}
\subsection{Preliminaries: Task Planning with PDDL}
In this work, tasks and actions are represented via \glsentryfull{pddl} \cite{fox2003pddl2}. Formally, a task planning problem is defined as a tuple $\langle \mathcal{O}, \mathcal{P}, s_0, \mathcal{A}, \tau \rangle$, where $\mathcal{O}$ represents all objects (e.g., robots, containers, and interactable objects), $\mathcal{P}$ denotes the set of predicates expressing relationships among these entities, $s_0$ is the initial state, and $\mathcal{A}$ is a set of parameterized actions. The \emph{task} $\tau$ is defined as a conjunction of predicates that specifies a subset of the state space in which the task is considered complete: the goal space $G_{\tau} \subseteq \mathcal{S}$.

\subsection{Sequential Task Planning in a Shared Environment}
We consider a shared environment with a set of robots $R$. Each robot $r\in R$ has its own \emph{responsibilities}: a set of tasks $\tau$ it may be assigned and their respective probability $P_r(\tau)$. We are given a sequence of $N=n\,|R|$ tasks $\tau$ containing $n$ sampled tasks for each robot $r$, which arrives one at a time. At each step $i$ in the sequence, a robot $r$ is randomly chosen and assigned a task $\tau_i \sim P_r(\tau)$ at random, drawn from its responsibilities.

The environment persists between tasks: the terminal state after completing one task becomes the starting state of the next. In notation, the state $s_{g_i}$ which is the terminal state of completing $\tau_i$ is also the initial state of completing the task $\tau_{i+1}$. It is the aim of each robot to plan in such a way that the cost over completing all tasks in the sequence is minimized. If the task sequence were known in advance, it would be possible to compute the minimum total-cost plan via the following equation.
\begin{equation} \label{eq:full-problem}
\begin{split}
s^*_{g_1}, \cdots\!,  s^*_{g_N} & = \!\!\!\!\!\! \underset{{s_{g_i}\in G_{\tau_i}} \forall i \in {1 \cdots n}}{\text{argmin}}\!\!\!\!\!\! \left[\substack{V^*_{s_{g_1}}(s_0)+V^*_{s_{g_2}}(s_{g_1}) + \cdots + V^*_{s_{g_n}}(s_{g_{n-1}})}\right].\\
\end{split}
\end{equation} where $V^*_{s_{g_{i}}}(s_{g_{i-1}})$ is the cost of reaching the terminal state of task $\tau_i$ $s_{g_{i}}$ from $s_{g_{i-1}}$ considering the complete sequence of tasks ahead.

Since the robot will not, in general, know all its own tasks and those of others in advance, it instead seeks a plan to complete the immediate task that minimizes \emph{expected total cost}, with future tasks drawn from each robot's responsibilities. Computing this expected total cost in general is computationally infeasible as the space of possible future task sequences grows combinatorially with the number of robots and breadth of their responsibilities which motivates our approximation approach described in Sec.~\ref{sec:method}.



%% file: Section/Approach.tex
\section{Courteous Anticipation in a Persistent Shared Environment}
With the problem formulation of Sec.~\ref{sec:problem} in hand, we develop our courteous anticipatory planning objective, grounding it in single-robot anticipatory planning before presenting a formulation that reasons about the distinct capabilities and responsibilities of all robots sharing the environment.

Recent work~\cite{dhakal2023anticipatory, talukder2025} proposes single-robot \emph{anticipatory planning}: a planning objective that minimizes the joint cost of the current task and the expected cost associated with future tasks that robot may be assigned next. Given the current task $\tau_c$ with starting state $s_0$ and defining the terminal state of the plan $\pi$ as $s'_g \equiv \textsc{Tail}(\pi) \in G_{\tau_c}$ this objective is:
\begin{equation}\label{eq:antplan}
\begin{split}
s^*_g & = \underset{s'_g\in G_{\tau_c}}{\text{argmin}} \Big[ V_{s'_g}(s_0) + \sum_{\tau'} P(\tau') V_{\tau'}(s'_g) \Big] \\ 
& = \underset{s'_g\in G_{\tau_c}}{\text{argmin}} \Bigg[V_{s'_g}(s_0) + \apcost{} (s'_g)\Bigg].
\end{split}
\end{equation}
where $V_{s'_g}(s_0)$ is the cost to reach goal state $s'_g \in G_{\tau_c}$ from $s_0$ and $V_{\tau'}(s'_g)$ is the cost of completing a future task $\tau'$ from state $s'_g$. $\apcost{}(s'_g)$ is the \emph{anticipatory cost}: shorthand for the expected cost to complete a follow-up task from state $s'_g$.

While effective for a single robot, this formulation remains \emph{selfish} in shared environments. Absent any consideration of how robot's current actions impact \emph{other} robots that share the environment, the robot may behave \emph{selfishly}, creating downstream side-effects that increase total cost over task sequences as illustrated in (Fig.~\ref{fig:example}). 
So, minimizing Eq.~\eqref{eq:antplan} alone cannot resolve this, as it is inconsiderate to the capabilities and responsibilities of others. 
To capture the impact of planning decisions on all robots sharing the 
environment, we propose \emph{courteous anticipation}. Generalizing 
Eq.~\eqref{eq:antplan}, each robot $r \in R$ draws its own next task 
$\tau'_r \sim P_r(\tau'_r)$ from its responsibilities, and the acting robot 
selects its terminal state to jointly minimize its immediate cost and the 
expected cost this induces across all robots' next tasks:

\begin{equation}\label{eq:courtesy}
\begin{split}
s^*_g &= \underset{s'_g \in G_{\tau_c}}{\text{argmin}}\, 
\Bigg[ V_{s'_g}(s_0) + \sum_{r \in R} V^r_{\textsc{A.P.}}(s'_g) \Bigg] \\ &=
\underset{s'_g \in G_{\tau_c}}{\text{argmin}}\, 
\Bigg[ V_{s'_g}(s_0) + V^R_{\textsc{A.P.}}(s'_g) \Bigg] 
\end{split}
\end{equation}

Unlike Eq.~\eqref{eq:antplan}, which considers only the acting robot's own 
next task, Eq.~\eqref{eq:courtesy} sums this anticipatory cost over 
every robot sharing the environment, capturing distinct responsibility 
distributions and structural constraints across robots and altering the 
terminal-state ranking. Exactly evaluating $V^R_{\textsc{A.P.}}(s'_g)$ 
requires solving every task in robot $r$'s responsibility set with a 
task planner for each candidate terminal state; repeated across all 
robots and every step of a lengthy task sequence, this becomes 
computationally expensive as responsibilities and number of robots grow (Sec.\ref{sec:exp:rest}). We instead estimate each $V^r_{\textsc{A.P.}}(s'_g)$ with a 
learned model trained specifically on robot $r$'s responsibilities implicitly capturing how the terminal state $s'_g$ interacts with that robot's capabilities, such as objects placed in locations inaccessible to robot $r$, or left in states it cannot change, will incur higher estimated costs, encoding the capability constraints. This decomposition remains tractable while capturing the primary source of cross-robot side effects: the physical configurations left by the acting robot.

This per-robot decomposition further supports \emph{modular deployment}: 
because each estimator is trained independently on a single robot's 
responsibilities, adding a robot to the shared environment requires only 
training that robot's own estimator, without retraining others already 
deployed, a property we validate empirically in Sec.\ref{sec:exp:rest} and contrast against a monolithic, jointly-trained alternative.

%% file: Section/Methodology.tex
\section{Courteous Anticipatory Planning} \label{sec:method}
\begin{figure}
    \vspace{0.6em}
    \centering
    \includegraphics[width=8.5cm]{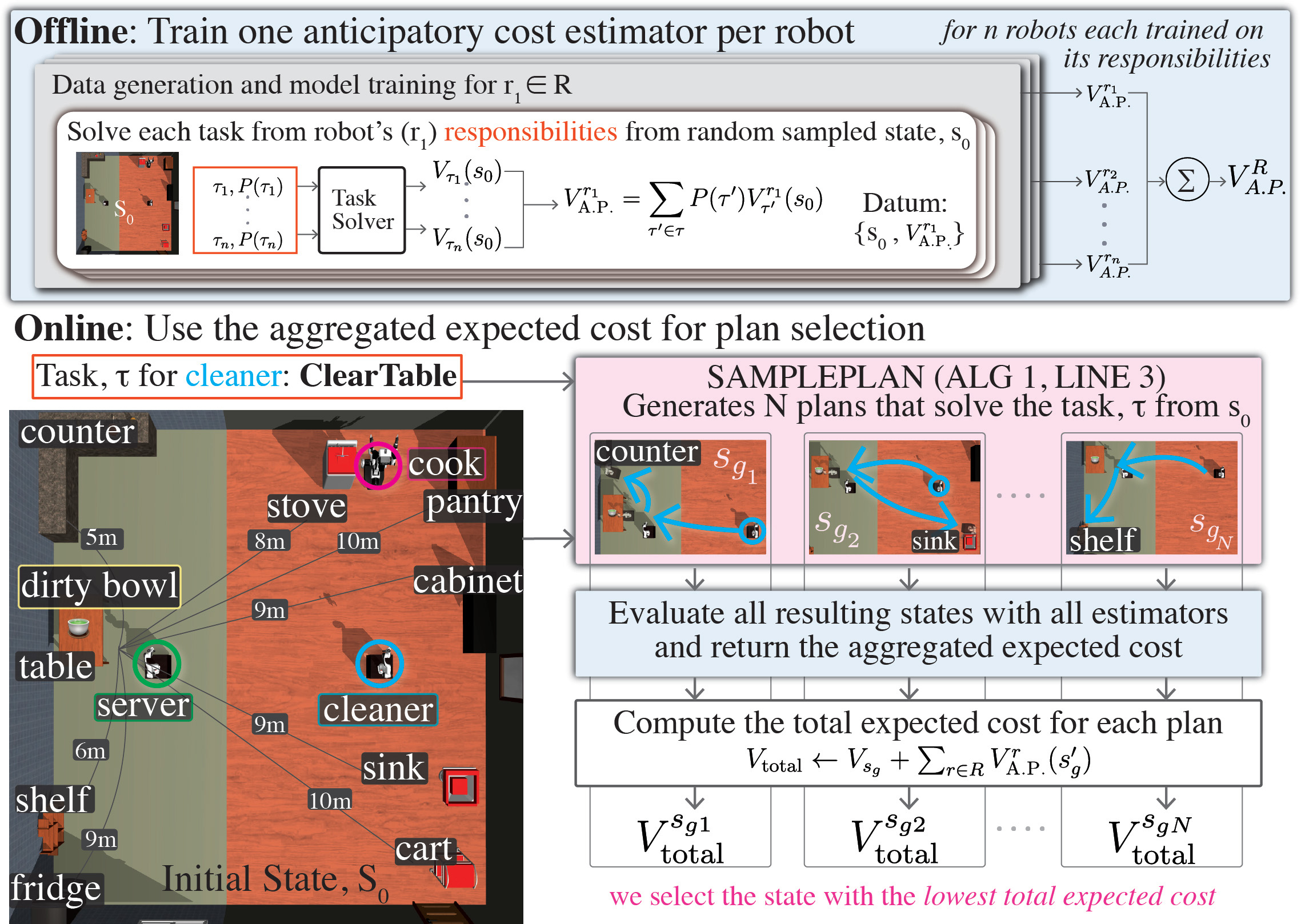}
    \caption{\textbf{Courteous Planning:} Our courteous anticipatory planning generates a set of candidate plans and selects from them the plan that jointly minimizes the cost of immediate plan and the expected future cost aggregated for all robots.}\label{fig:example-ap-vs-mp}
\end{figure}

In this section, we describe our courteous anticipatory planner Alg.~\ref{alg:cap} which proposes candidate plans that achieve the current task and, from the resulting set, selects the plan that minimizes the joint cost of the immediate plan and expected future cost aggregated from each robot, relying on a per-robot learned model to estimate the anticipatory cost for each. We detail the plan-sampling procedure in Sec.~\ref{sec:search} (Alg.~\ref{alg:cap}, Line~3) and the per-robot anticipatory cost estimator in Sec.~\ref{sec:learning} (Alg.~\ref{alg:cap}, Line~5).
Fig.~\ref{fig:example-ap-vs-mp} includes a schematic overview of our approach.

\begin{algorithm}[h]
\caption{Courteous Anticipatory Planning}\label{alg:cap}
\LinesNumbered
\DontPrintSemicolon
\footnotesize
\KwIn{$s_0, \tau$}
\KwOut{$\pi^*$}
$\pi^* \gets \emptyset$,$V^*_{\text{total}} \gets \infty$\;
\For{$i \in \{1,2,...,N\}$}{
    $\pi \gets \textsc{SamplePlan}(s_0, \tau, r_{\text{act}})$\;
    $s'_g, V_{s'_g} \gets \textsc{Tail}(\pi)$\;
    $V_{\text{total}} \gets V_{s'_g} + \sum_{r \in R} \textsc{APCostEstimator}_r(s'_g)$\;
    \If{$V_{\text{total}} \leq V^*_{\text{total}}$}{
        $V^*_{\text{total}} \gets V_{\text{total}}$\;
        $\pi^* \gets \pi$\;
    }
}
\Return{$\pi^*$}\;
\end{algorithm}

\subsection{Generating Candidate Plans}\label{sec:search}
In large environments the number of plans satisfying a task can grow exponentially, quickly making planning intractable. We adopt a focused sampling procedure~\cite{talukder2025}, which augments tasks with placement predicates and object state predicates—specifying where moved objects should be placed and in what state they should be left—to generate candidate plans that are structurally distinct in ways that matter for downstream cost, rejecting any that fail to accomplish the original task. This property is essential for courteous anticipation, where the acting robot must evaluate how different object placements and states affect the future costs of all robots sharing the environment. In a \texttt{ClearTable} task, for instance, the cleaner has discretion over where a dirty bowl is placed and whether it is left clean or dirty, and each choice has different downstream consequences for other robots (e.g., a dirty bowl left on the counter is unusable by the cook until the cleaner washes it). We use FastDownward~\cite{helmert2006fast} with \texttt{ff-astar}~\cite{Hoffmann2001FFTF} to solve each proxy task, producing a set of candidate plans from which we select the one that minimizes Eq.~(\ref{eq:courtesy}).

\subsection{Per-Robot Anticipatory Cost Estimation}\label{sec:learning}
The full joint evaluation needed for the exact computation of expected future cost via Eq.~(\ref{eq:full-problem}) is combinatorially expensive. Instead, our courteous anticipatory cost formulation of Eq.~\eqref{eq:courtesy} factors the expected future cost to separately reason about the expected impact of the immediate plan on each robot's next task, drawn from their respective responsibilities.
Thus, we train a separate anticipatory cost estimator for each robot independently, where the expected future cost is defined by that robot's responsibilities: a distribution over what tasks it may be assigned. At deployment, the planner queries all per-robot estimators and aggregates their outputs to approximate the expected future cost across all robots sharing the environment. Because each estimator is trained independently on a single robot's responsibilities and implicitly captures the robot's structural constraints, adding a robot to the shared environment requires only training that robot's own estimator, without retraining of existing ones, a property we validate empirically in Sec.~\ref{sec:exp:rest}.

\subsubsection{Data Generation}
We generate training data from held-out training-time environments. It is assumed that each robot has direct access to its responsibilities, which specifies both the tasks it may be assigned and their relative likelihoods.
Starting from the initial environment state, we solve tasks from these responsibilities using FastDownward~\cite{helmert2006fast} with an informed search (\texttt{ff-astar}~\cite{Hoffmann2001FFTF}) and use the resulting plan costs to compute the anticipatory cost, $V^r_{\textsc{A.P.}}(s'_g)$, for that robot $r$. We represent the world as a sparse scene graph: nodes are entities with semantic and geometric attributes and edges encode spatial or logical relations. This graph defines our \gls{gnn}, which estimates the expected cost of a state. Because this data is generated offline, we can afford more computation to approximate the expectation accurately; at deployment, we avoid rollouts and only query the learned estimators, keeping online planning fast.

\subsubsection{Per-Robot Anticipatory Cost Estimators}
Each per-robot estimator is a graph neural network (GNN) operating over the scene graph representation, trained independently on a single robot's responsibilities. We use four GINConv layers followed by batch normalization and leaky ReLU activations. Node features are aggregated into a graph-level representation via both mean and sum pooling before a final linear layer produces the anticipatory cost estimate.
At planning time, the planner queries all per-robot estimators over the candidate terminal states produced by focused sampling, aggregates their outputs across all robots, and selects the plan whose terminal state minimizes Eq.~(\ref{eq:courtesy}).

%% file: Section/Experiment.tex
\section{Experiments and Results} \label{sec:experiments}
We evaluate courteous anticipatory planning in two simulated environments of increasing complexity. In the home setting, courteous anticipation improves performance even without capability differences. In the restaurant environment, where robots have distinct capabilities and responsibilities, results validate that capability differences amplify the value of courteous anticipation.

\begin{figure}
    \vspace{0.6em}
    \includegraphics[width=8.5cm]{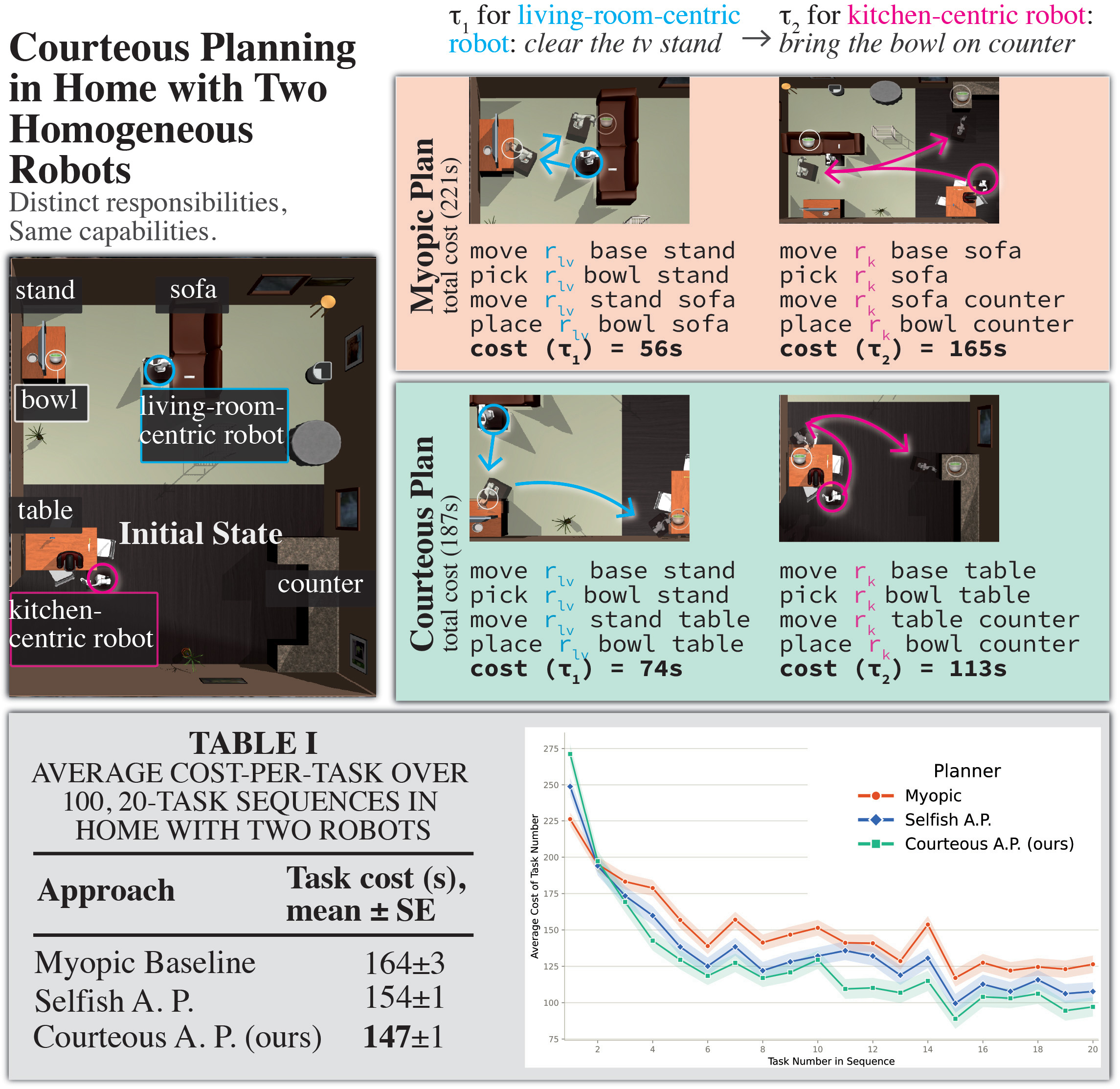}
    \centering
    \caption{\textbf{Courteous Planning in Home Environments:} An example from our results in which the living-room-centric robot is first tasked to clear the tv stand and then, later, the kitchen-centric robot is assigned a task of its own. By clearing the tv stand more courteously, our courteous living-room-centric robot makes it easier for the kitchen-centric robot later. \emph{See Sec.~\ref{sec:exp:homes} for details.}
    }
    \label{fig:procthor-example}
    \vspace{-1.5em}
\end{figure}

Both domains present large, combinatorially rich planning problems: our home
environments cap at 50 objects across 2-room ProcTHOR~\cite{procthor} houses,
yielding ground action spaces of up to $\sim$4,760, while our restaurant
domain yields up to $\sim$5,730 ground actions despite fewer objects(25), owing
to its higher-arity, role-specific actions (\texttt{mix}, \texttt{serve}).
This scale poses a meaningful search challenge for anticipatory planning, where each candidate terminal state must be evaluated against every robot's responsibilities.

In all experiments, planning cost reflects the physical time required to execute each action: only movement actions incur a dynamic cost proportional to travel distance, while other actions carry fixed unit costs. The specific manipulation actions vary by environment, with home environments involving \texttt{pick} and \texttt{place} and the restaurant environment additionally including \texttt{cook}, \texttt{wash}, \texttt{wipe}, \texttt{restock}, \texttt{serve}, \texttt{pour} and \texttt{request-help}, all evaluated under the same cost structure.

We compare against three baselines in both environments:


\begin{LaTeXdescription}\label{planning:approach}
  \item[\colorunderline{myopicColor}{Myopic Baseline}] Classical planning via FastDownward
  \item[\colorunderline{selfishColor}{Selfish Anticipatory Planning (Selfish A.P.)}] Planning augmented with the acting robot's own anticipatory cost estimator via Eq.~\eqref{eq:antplan}, reducing cost for the acting robot's future tasks but not for others sharing the environment.
  \item[\colorunderline{courteousColor}{Courteous Anticipatory Planning (Courteous A.P.)}] Our approach, planning augmented with aggregated per-robot anticipatory cost estimates via Eq.~\eqref{eq:courtesy}, accounting for all robots sharing the environment.
  \item[\colordashunderline{courteousJointColor}{Courteous A.P. (Joint)}] An ablation of our approach that replaces the per-robot estimators with a single estimator trained jointly on all robots' responsibilities, requiring full retraining whenever a robot is added to or removed from the environment.
  \vspace{-0.1em}
\end{LaTeXdescription}

\subsection{Courteous Anticipation from Responsibilities Alone}\label{sec:exp:homes}
\textbf{The Home Environment}: Home environments are curated from \gls{procthor}~\cite{procthor} and include two robots with similar capabilities but distinct responsibilities: rearrangement-style tasks curated from the \gls{alfred}~\cite{shridhar2020alfred} task planning benchmark. 
Because both robots can physically interact with any asset (object and container) in the scene and perform the same actions, improvements arise solely from reasoning about distinct responsibilities and shared surfaces.
The kitchen-centric robot's responsibilities include clearing the counter, putting food in the fridge, and organizing other kitchen items: e.g., placing mugs and bowls on shelves. The living-room-centric robot's responsibilities include cleaning containers and organizing living room objects such as arranging pillows on the sofa or clearing the desk. Though robots are assigned tasks independently, their shared environment and surfaces create opportunities for courteous anticipation to reduce downstream friction, without explicitly encoding awareness of others' responsibilities.

\textbf{Results}: We include a specific example that illustrates this behavior in Fig.~\ref{fig:procthor-example}. First, the living-room organizer is tasked to \texttt{ClearStand}, requiring it to remove all objects from the tv stand (here, only a bowl). The myopic planner moves the bowl to the nearby sofa without considering downstream consequences. While our courteous planner places the bowl on a slightly further dining table which is near the kitchen area, anticipating the kitchen organizer's future responsibilities. So when the kitchen organizer is subsequently tasked to bring the bowl to counter, our approach incurs lower cost.

We conduct experiments consisting of 100 sequences of 20 tasks each (10 per robot, drawn from that robot's responsibilities), and report statistics in Table 1 (Fig.~\ref{fig:procthor-example}).
Courteous anticipation reduces cumulative planning cost by 10.4\% versus the myopic planner and 4.5\% versus the selfish anticipatory planner. As shown in the plot of average cost over the sequence in Fig.~\ref{fig:procthor-example}, because the \gls{procthor} are initialized with objects placed in somewhat random locations throughout the homes, all planners tend towards lower-expected-cost configurations over time, an observation shared by Talukder et al.~\cite{talukder2025}. However, our more courteous robots are able to leave objects in locations more favorable for \emph{all} robots and improve over both baselines, made possible by their anticipation of how their actions will impact future tasks that both it and other robots may later be assigned. This improvement, achieved with similar capabilities and only distinct responsibilities, establishes a conservative lower bound on the value of courteous anticipation. As we show next, introducing distinct capabilities amplifies these gains substantially.

\begin{figure*}[t]
    \centering
    \vspace{0.6em}
    \includegraphics[width=0.95\textwidth]{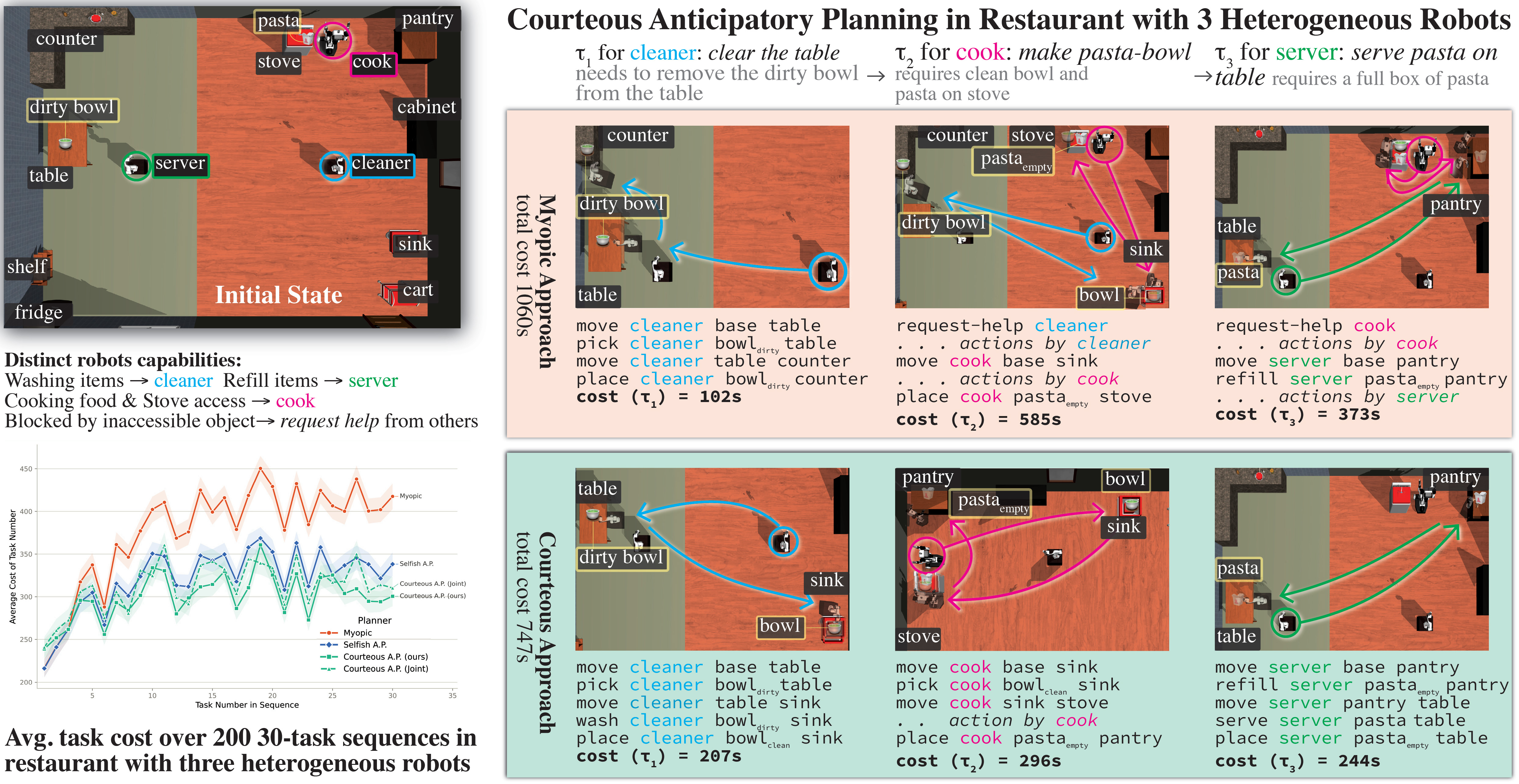}
    \vspace{-1em}
    \caption{\textbf{A Courteous Anticipation Sequence in the Restaurant Environment:}
    An example sequence of three tasks that highlights the value of courtesy in our restaurant domain. Preemptive courteous action by the Cleaner during task 1 pays considerable dividends over time, improving the performance of both the Cook and Server later on as compared to the myopic planner, which has no such foresight.
    (bottom-left) Plots the average per-task performance over the task sequences in 3-robot settings.
    }\label{fig:exp-restaurant}
    \vspace{-0.6em}
\end{figure*}

\subsection{Distinct Capabilities Amplify Courteous Anticipation}\label{sec:exp:rest}
\textbf{The Restaurant Environment}: To evaluate the full setting motivating our approach, we develop a restaurant-inspired benchmark environment with three heterogeneous robots that differ in physical capabilities and responsibilities, making one robot's decisions directly consequential for what others can accomplish.
The \emph{Cook} is tall and alone can safely access the stove. The \emph{Server} is also taller and alone can reach and serve items on dining table. The \emph{Cleaner} alone is equipped to wash cookware. Consistent with their roles, the Cook's responsibilities include \texttt{MakingFood} objectives, the Cleaner's responsibilities include \texttt{CleanItem} and \texttt{ClearContainers} objectives, and the Server's responsibilities include \texttt{ServeFood} and \texttt{RestockFood} objectives.


We use two-room environments from \gls{procthor} as a base, populating them with domain-specific objects and actions suitable for food preparation and kitchen maintenance, from which complex multi-step tasks are constructed for each robot consistent with their roles.

Sometimes a robot cannot complete a task on its own because an object has been placed in a location it cannot physically access, or because an object is in a state it cannot change, for example a dirty pan that only the \emph{Cleaner} can wash, or an empty pasta box on dining table that only the \emph{Server} can fetch and restock. For this purpose, we add a \texttt{request-help} action that activates an otherwise \emph{idle} robot—one not currently assigned a task—to satisfy the precondition on behalf of the acting robot. Because each robot has at least one exclusive action, such inaccessible terminal state arise when robots place objects without regard for others' structural constraints. Invoking \texttt{request-help} accumulates an additional fixed cost of 100 sec, reflecting the real cost of interrupting an idle robot, incentivizing courteous behavior: a robot that reasons about what others can do and what they are likely to be assigned will leave objects in locations and states accessible to others, minimizing the need for costly assistance. We treat this cost as a design parameter of our benchmark and 
evaluate its impact on planner performance.

\setcounter{table}{1}
\begin{table}[t]
\caption{Per-robot results and improvements.}
\vspace{-0.6em}
\label{tab:results-per-robot}
\centering
\setlength{\tabcolsep}{2.8pt}
\renewcommand{\arraystretch}{1.15}
\resizebox{\columnwidth}{!}{%
\begin{threeparttable}
\begin{tabular}{ccc >{\columncolor{colMyopic}}c >{\columncolor{colSelfAP}}c >{\columncolor{colJointAP}}c >{\columncolor{colOursAP}}c >
{\columncolor{colMyopic}}c >{\columncolor{colSelfAP}}c >{\columncolor{colJointAP}}c >{\columncolor{colOursAP}}c
}
\toprule
\multicolumn{3}{c}{Robots Present} & \multicolumn{4}{c}{\makecell{Task cost (s), $mean \pm SE$}} & \multicolumn{3}{c}{\makecell{(\%) Improvement}} \\
\cmidrule(lr){1-3}\cmidrule(lr){4-7}\cmidrule(lr){8-10}
\makecell{Cook} & \makecell{Cleaner} & \makecell{Server} & \cellcolor{colDef}\makecell{Myopic} & \cellcolor{colDef}\makecell{Selfish \\ A.P.} & \cellcolor{colDef}\makecell{Court. A.P. \\ (joint)} & \cellcolor{colDef}\makecell{Court. A.P.\\ (ours)}  & \cellcolor{colDef}\makecell{vs Myopic.} & \cellcolor{colDef}\makecell{vs \\ Selfish A.P.} & \cellcolor{colDef}\makecell{vs \\ Court. A.P. (joint)} \\
\midrule
\checkmark & \graydash{}         & \graydash{}          & 510$\pm$9 & \textbf{456$\pm$9}\tnote{*} & \textbf{456$\pm$9}\tnote{*} & \textbf{456$\pm$9}\tnote{*} & 10.5$\pm$1.1\% & N/A\tnote{*} & N/A\tnote{*} \\
\graydash{}          & \checkmark & \graydash{}          & 100$\pm$4 & \textbf{99$\pm$4}\tnote{*} & \textbf{99$\pm$4}\tnote{*} & \textbf{99$\pm$4}\tnote{*} & 1.0$\pm$0.3\% & N/A\tnote{*} & N/A\tnote{*} \\
\graydash{}          & \graydash{}         & \checkmark & 391$\pm$7 & \textbf{372$\pm$7}\tnote{*} & \textbf{372$\pm$7}\tnote{*} & \textbf{372$\pm$7}\tnote{*} & 4.8$\pm$0.6\% & N/A\tnote{*} & N/A\tnote{*} \\[4px]
\checkmark & \checkmark & \graydash{}          & 377$\pm$2 & 324$\pm$3 & 319$\pm$3 & \textbf{318$\pm$3}          & 15.6$\pm$0.7\% & 1.6$\pm$0.5\% & 0.3$\pm$0.6\% \\
\graydash{}          & \checkmark & \checkmark & 290$\pm$3 & 277$\pm$3 & 275$\pm$3 & \textbf{269$\pm$3}          & 7.1$\pm$0.5\% & 2.8$\pm$0.5\% & 2.3$\pm$0.5\% \\
\checkmark & \graydash{}         & \checkmark & 601$\pm$3 & 532$\pm$3 & 528$\pm$4 & \textbf{507$\pm$3}          & 15.7$\pm$0.4\% & 4.8$\pm$0.6\% & 4.0$\pm$0.5\% \\[4px]
\checkmark & \checkmark & \checkmark & 377$\pm$3 & 321$\pm$3 & 313$\pm$3 & \textbf{299$\pm$4} & 20.5$\pm$0.5\% & 6.7$\pm$0.3\% & 4.1$\pm$0.3\% \\
\bottomrule
\end{tabular}%
\begin{tablenotes}[flushleft]
\footnotesize
\vspace{0.3em}
\item[] \parbox{\linewidth}{\emph{Note: Differences between each robot configuration mean costs are not straightforwardly comparable across rows as they represent distinct planning problems.}}\vspace{0.5em}
\item[] \parbox{\linewidth}{*For single-robot settings, there are no other agents to be courteous to, and so Courteous A.P. (joint), Courteous A.P. (ours), and Selfish A.P. planning yield identical results.}
\end{tablenotes}
\end{threeparttable}%
}
\vspace{-2em}
\end{table}

\textbf{Results}: 
We highlight an illustrative example in Fig.~\ref{fig:exp-restaurant} that shows the full three-robot restaurant, populated by the Cook, Server, and Cleaner.
When the Cleaner is tasked with \texttt{ClearTable}, our courteous anticipatory planner washes the dirty bowl at the sink and leaves it there; a slightly more costly plan for the Cleaner, but one that benefits both the Cook and Server later. By contrast, both the myopic 
and selfish planners leave the bowl dirty on a shared surface, 
obstructing it and requiring the Cook or Server to invoke costly 
\texttt{help} from the Cleaner when later assigned tasks of their own.

We evaluate performance beginning with a single robot, progressively adding robots to the shared environment, first two-robot subsets and then the full three-robot team, to showcase how courteous behavior evolves as robots are added or removed and to validate the modularity of our independently trained estimators. Across all configurations, we evaluate performance across 200 sequences of $10N$ tasks, where $N$ is the number of robots present: i.e., that a 3-robot configuration would be given a total of 30 tasks. 

Our statistical results in Table~\ref{tab:results-per-robot} shows that our courteous, factored anticipatory planning approach consistently outperforms myopic planning, selfish anticipatory planning, and a monolithic joint-courteous estimator across all configurations involving more than one robot. 
The gap over Myopic reflects the value of anticipation itself, while the smaller but consistent gaps over Selfish A.P. and Joint isolate the incremental benefit of courtesy and of factoring, respectively—both of which grow as more robots are included.

\textbf{Modularity Validation.} A key claim of our approach is that per-robot estimators trained independently can be composed without retraining when robots are added to or removed from the shared environment. Table~\ref{tab:results-per-robot} validates this directly: estimators trained in two-robot settings are reused unchanged in three-robot experiments with no joint retraining, confirming the plug-and-play property claimed in Sec. \ref{sec:learning}.
By contrast, Courteous A.P. (Joint) requires training a separate monolithic estimator for each configuration in Table~\ref{tab:results-per-robot}—seven trainings in total—since its single model is tied to a fixed composition and cannot be composed or reused. Despite this substantially higher training cost, Joint achieves worse plan quality to our factored approach, underscoring that factoring is not merely a modularity convenience but yields plans of better quality without the retraining burden. As expected, courteous and selfish planning yield identical behavior in the single-robot setting, confirming that courtesy emerges only when other robots are present.

\begin{figure}
  \centering
  \vspace{0.6em}
  \includegraphics[width=8.5cm]{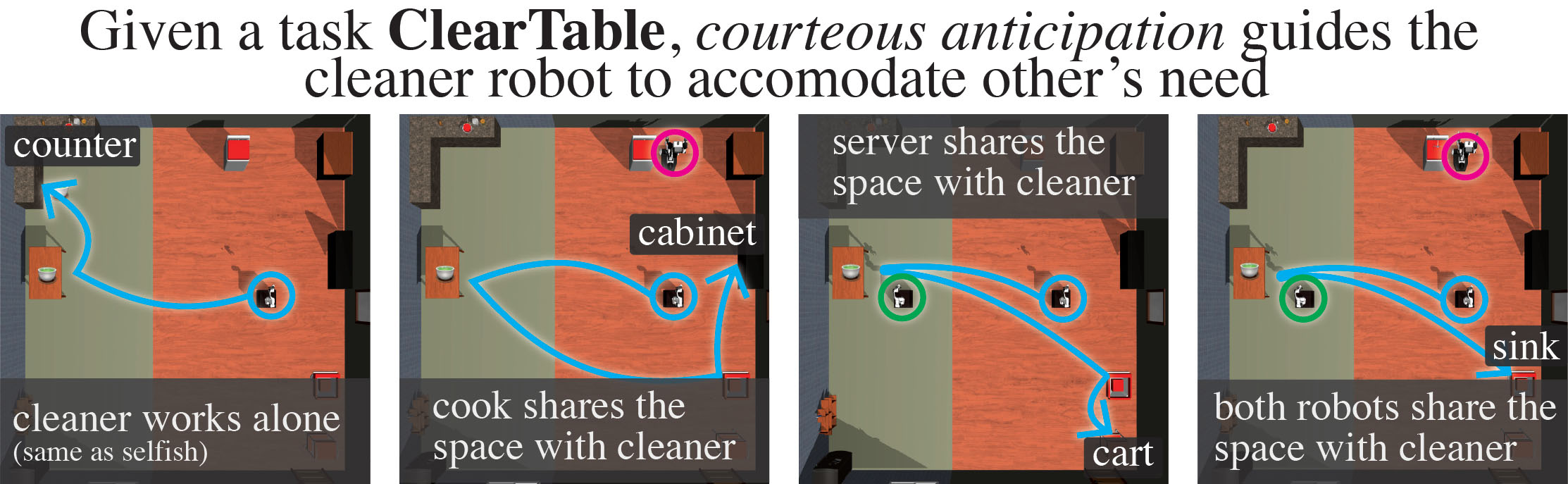}
  \caption{\textbf{Courteous Planning in our Restaurant Environment:} Here we show an example of how the addition of other robots in the scene prompts behavior changes in our courteous robot. \emph{See Sec.~\ref{sec:exp:rest} for details.}}\label{fig:cook-behave}
  \vspace{-1.6em}
\end{figure}

\textbf{Distinct Capabilities Drive Courteous Behavior.} The robot-subset experiments, summarized in Fig.~\ref{fig:cook-behave}, reveal how distinct capabilities shape courteous behavior. When the Cleaner operates alone on \texttt{ClearTable}, courteous and selfish planning yields the same result (moves the dirty bowl to nearest counter). When the Cook is added, the courteous Cleaner additionally washes the bowl and moves it to the cabinet, since it knows the Cook might need a clean bowl and keeping it near the stove would reduce its cost. When the Server is added, the courteous Cleaner places clean bowl on cart; close to Server. When all three robots are present, the Cleaner places the clean bowl on sink; good for all robots. This progressive behavioral change, driven by distinct capabilities, directly validates the core claim of our introduction: a robot must reason about what others can do, not only what they will do. Single-robot experiments are included for completeness, though when only the single robot is included both selfish and courteous anticipatory planning yield identical behavior.

\textbf{Computational Cost of Learned vs.\ Exact Estimation.} 
As established in Sec. \ref{sec:problem}, exactly evaluating Eq. \eqref{eq:full-problem} is intractable, precluding a fully centralized joint-planning baseline; we instead treat exact per-robot cost computation as the strongest tractable baseline against which to validate our learned estimator.
A separate question is whether the per-robot anticipatory costs $V^r_{\textsc{A.P.}}$ 
must be learned, or could instead be computed exactly at planning time. We 
compare against \textbf{Courteous A.P. (exact)}, which solves every 
task in each robot's responsibility set directly with FastDownward for 
every candidate terminal state, rather than querying a learned estimator 
(Table~\ref{tab:help-ablation}). Learning matches exact's plan quality while reducing planning time by 7.0$\times$, confirming our learned estimators preserve quality while making courteous 
anticipation practical to deploy.

\begin{figure}
    \vspace{0.6em}
    \includegraphics[width=8.5cm]{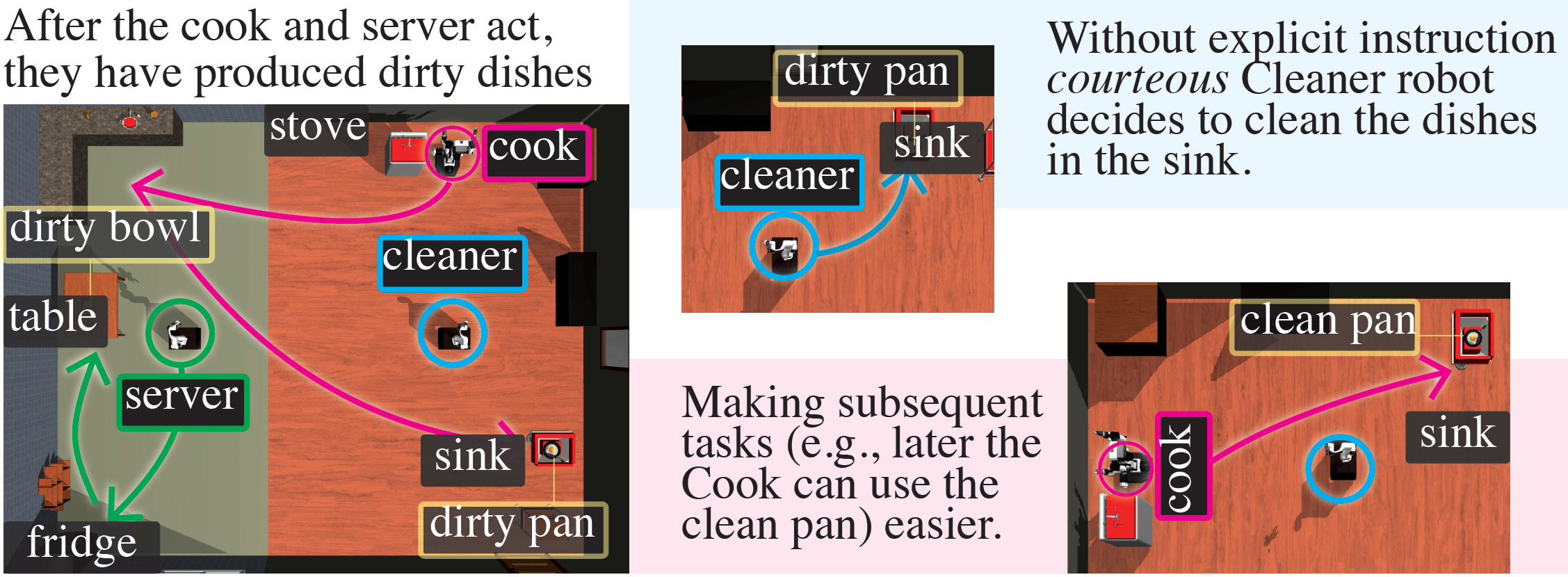}
    \centering
    \caption{\textbf{Proactivity via Courteous Anticipation:} In this scenario in our Restaurant domain, the Cleaner robot is not given explicit tasks and instead provided a small time budget to act during downtime when neither the Cook or Server are operating. Informed by its \emph{courteous anticipation}, the Cleaner proactively cleans dirty dishes in the sink, action that later benefits the Cook. \emph{See Sec.~\ref{sec:exp:rest} for details.}}\label{fig:idle-preparation}
    \vspace{-1em}
\end{figure}

\begin{table}[t]
\caption{Effect of request help action cost on Avg. Task cost}
\label{tab:help-ablation}
\centering
\setlength{\tabcolsep}{2.5pt}
\renewcommand{\arraystretch}{1.15}
\resizebox{\columnwidth}{!}{%
\begin{threeparttable}
\begin{tabular}{l >{\columncolor{colMyopic}}c >{\columncolor{colSelfAP}}c >{\columncolor{colJointAP}}c >{\columncolor{colOursAP}}c >
{\columncolor{colMyopic}}c >{\columncolor{colSelfAP}}c >{\columncolor{colJointAP}}c >{\columncolor{colOursAP}}c
}
\toprule
& \multicolumn{4}{c}{\makecell{Avg. Task Cost (s), $mean \pm SE$}} & \multicolumn{4}{c}{\makecell{\textit{Planner Time (s), $mean \pm SE$}}} \\
\cmidrule(lr){2-5}\cmidrule(lr){6-9}
\makecell{Help\\level} & \cellcolor{colDef}\makecell{Myopic} & 
\cellcolor{colDef}\makecell{Selfish\\A.P.} & \cellcolor{colDef}\makecell{Courteous\\ A.P.\\ (ours)} 
&\cellcolor{colDef} \makecell{Courteous\\ A.P. \\ (exact)} 
& \cellcolor{colDef}\makecell{Myopic} & 
\cellcolor{colDef}\makecell{Selfish\\A.P.} & \cellcolor{colDef}\makecell{Courteous \\ A.P. \\(ours)} & \cellcolor{colDef}\makecell{Courteous\\ A.P. \\ (exact)} \\
\midrule
0   & 272$\pm$5.7 & 244$\pm$7.2 & 246$\pm$7.0 & 245$\pm$6.6 & \textit{0.1$\pm$0} & \textit{28$\pm$1.5} & \textit{31$\pm$1.7} & \textit{162$\pm$4.5} \\
100 & 336$\pm$8.6 & 316$\pm$9.8 & 283$\pm$7.8 & 283$\pm$7.9 & \textit{0.1$\pm$0} & \textit{48$\pm$8.9} & \textit{30$\pm$1.0} & \textit{173$\pm$4.7} \\
200 & 408$\pm$8.4 & 356$\pm$12.1 & 337$\pm$10.2 & 323$\pm$10.5 & \textit{0.1$\pm$0} & \textit{60$\pm$9.0} & \textit{17$\pm$0.5} & \textit{170$\pm$5.4} \\
300 & 453$\pm$12.7 & 399$\pm$13.8 & 352$\pm$12.7 & 337$\pm$11.0 & \textit{0.1$\pm$0} & \textit{22$\pm$1.1} & \textit{17$\pm$0.8} & \textit{156$\pm$4.5} \\
\bottomrule
\end{tabular}%
\begin{tablenotes}[flushleft]
\footnotesize\vspace{0.3em}
\item[] \parbox{\linewidth}{\emph{Note}: \textbf{Task Cost}: the average cost to complete a task by robots, reflecting plan \emph{quality}. \textbf{Planner Time}: the wall-clock time the planner takes to compute a plan per task, reflecting computational \emph{cost}. Courteous A.P. (ours) is our factored estimator; Courteous A.P. (exact) is the exact expected-cost baseline. This ablation uses a reduced \textbf{15-task} sequence (vs. 30 tasks in Sec.~\ref{sec:exp:rest}) due to the computational cost of the exact expected-cost baseline over long horizons; absolute values are not directly comparable to Table.~\ref{tab:results-per-robot}.}
\end{tablenotes}
\end{threeparttable}%
}
\vspace{-2em}
\end{table}

\textbf{Sensitivity to the \texttt{request-help} Action Cost.} The 
\texttt{request-help} action's fixed cost (100s in our main experiments, 
Sec.~\ref{sec:experiments}) is a design parameter of our benchmark rather 
than an intrinsic property of courteous anticipation. 
Table~\ref{tab:help-ablation} reports average task cost as this parameter varies from 0 to 300s. 
Myopic's cost grows sharply (+66.5\%), since it has no mechanism to foresee that its terminal state will later require assistance and so pays this penalty repeatedly. 
Selfish A.P. degrades at a nearly identical rate (+63.5\%): although it anticipates its own future needs, it does not model how its actions affect \emph{other} robots' need for help, and so triggers avoidable \texttt{request-help} calls just as often as Myopic.
At $help=0$ selfish and both courteous variants perform comparably, since invoking help is free and modeling others' needs offers little benefit; the residual gap to myopic instead reflects wasted reactive steps taken to complete any task.
Both courteous variants grow markedly more slowly (+43.1\% ours, +37.6\% exact) as request-help cost rises, widening their relative advantage over Myopic from roughly 10\% at cost 0 to over 20\% at cost 300; confirming that this benefit scales with, rather than is an artifact of, our specific cost choice. 

\textbf{Proactivity via Courteous Anticipation}: 
We additionally conduct experiments in which one of the robots in our restaurant environment is given \emph{no explicit task} and instead given a limited time budget to proactively ready the environment during downtime when the other two robots are not assigned tasks. Our experiments augment the two-courteous-robot Cook-Server experiments from the previous section and additionally include a free Cleaning robot that, after every two tasks, is given a cost budget of 100 sec to act in the environment to reach the state that maximally reduces the aggregated expected future cost for the Cook and Server robots; we use simulated annealing for this search process, iteratively perturbing the state space as computation allows.

Made possible by our courteous anticipatory planning approach, the addition of the free courteous Cleaning robot reduces cost from 507s to 263s, a 48.1\% improvement. 
By contrast, a selfish or myopic robot would not take \emph{any} action while idle, unaware of how its actions could help improve the performance of other robots in the scene. Thus, our courteous anticipation approach helps to motivate effective behaviors not exhibited by other approaches, affording improvements in persistent shared environments.


%% file: Section/Conclusion.tex
\section{Limitations and Future Work}

We have presented courteous anticipatory planning, a framework for reducing cumulative cost in persistent shared environments by reasoning about what other robots can do and what they are likely to be assigned. Our factored formulation estimates per-robot expected future costs independently, avoiding combinatorial joint rollouts and supporting modular deployment without retraining existing estimators. Across two environments of increasing complexity, amplified gains in the restaurant setting directly validate that distinct capabilities amplifies the value of courteous anticipation.

Our approach assumes tasks arrive sequentially, that the environment persists between tasks, which is reasonable for many shared environments but does not capture all possible real-world situations. We acknowledge that this setting does not capture all multi-agent challenges: tasks may arrive concurrently, robots may act simultaneously, and the environment may change mid-execution due to other agents or interruptions. However, we note that courteous anticipatory planning in shared environments is a symbolic planning behavior problem rather than a multi-agent coordination or communication problem. The multi-agent complexity we address is not joint policy learning ~\cite{lowe2017maddpg, rashid2018qmix} or emergent coordination, but rather the challenge of planning with foresight about others, a challenge unaddressed by existing anticipatory planning methods~\cite{dhakal2023anticipatory, talukder2025, task-anticipation}.

In our future work, we plan to extend courteous anticipatory planning to online settings, where multiple robots may act simultaneously, and temporal planning over overlapping schedules is necessary. We further plan to deploy and evaluate this approach on physical multi-robot systems, to validate online courteous task planning under real-world sensing, timing, and execution constraints.


%% file: references.bib
@inproceedings{plaku2010sampling,
  title={Sampling-based motion and symbolic action planning with geometric and differential constraints},
  author={Plaku, Erion and Hager, Gregory D},
  booktitle={IEEE International Conference on Robotics and Automation (ICRA)},
  year={2010},
}

@inproceedings{garrett2020pddlstream,
  title={{PDDLStream}: Integrating symbolic planners and blackbox samplers via optimistic adaptive planning},
  author={Garrett, Caelan Reed and Lozano-P{\'e}rez, Tom{\'a}s and Kaelbling, Leslie Pack},
  booktitle={Proceedings of the International Conference on Automated Planning and Scheduling},
  year={2020}
}

@inproceedings{kaelbling2011hierarchical,
  title={Hierarchical task and motion planning in the now.},
  booktitle={IEEE international conference on robotics and automation (ICRA)},
  author={Kaelbling, Leslie Pack and Lozano-P{\'e}rez, Tom{\'a}s},
  year={2011},
}

@inproceedings{srivastava2014combined,
  title={Combined task and motion planning through an extensible planner-independent interface layer},
  author={Srivastava, Siddharth and Fang, Eugene and Riano, Lorenzo and Chitnis, Rohan and Russell, Stuart and Abbeel, Pieter},
  booktitle={2014 IEEE international conference on robotics and automation (ICRA)},
  year={2014},
}

@article{kim2019learning,
  title={Learning to guide task and motion planning using score-space representation},
  author={Kim, Beomjoon and Wang, Zi and Kaelbling, Leslie Pack and Lozano-P{\'e}rez, Tom{\'a}s},
  journal={The International Journal of Robotics Research},
  year={2019},
  publisher={SAGE Publications Sage UK: London, England}
}

@inproceedings{chitnis2016guided,
  title={Guided search for task and motion plans using learned heuristics},
  author={Chitnis, Rohan and Hadfield-Menell, Dylan and Gupta, Abhishek and Srivastava, Siddharth and Groshev, Edward and Lin, Christopher and Abbeel, Pieter},
  booktitle={2016 IEEE International Conference on Robotics and Automation (ICRA)},
  year={2016},
}

@inproceedings{kim2020learning,
  title={Learning value functions with relational state representations for guiding task-and-motion planning},
  author={Kim, Beomjoon and Shimanuki, Luke},
  booktitle={Conference on Robot Learning},
  year={2020},
}

@inproceedings{dantam2016incremental,
  title={Incremental task and motion planning: A constraint-based approach},
  author={Dantam, Neil T and Kingston, Zachary K and Chaudhuri, Swarat and Kavraki, Lydia E},
  booktitle={Robotics: Science and Systems},
  year={2016},
}

@inproceedings{talukder2025,
  title={Anticipatory Planning for Performant Long-Lived Robot in Large-Scale Home-Like Environments},
  author =     {Talukder, Md Ridwan Hossain and Arnob, Raihan Islam and Stein, Gregory J.},
  booktitle =  {International Conference on Robotics and Automation (ICRA)},
  year =       {2025},
}

@article{fox2003pddl2,
  title={{PDDL2.1}: An extension to {PDDL} for expressing temporal planning domains},
  author={Fox, Maria and Long, Derek},
  journal={Journal of Artificial Intelligence Research},
  year={2003}
}

@article{helmert2006fast,
  title={The {Fast Downward} planning system},
  author={Helmert, Malte},
  journal={Journal of Artificial Intelligence Research},
  year={2006}
}

@article{Hoffmann2001FFTF,
author = {Hoffmann, J\"{o}rg and Nebel, Bernhard},
title = {The {FF} Planning System: Fast Plan Generation through Heuristic Search},
year = {2001},
issue_date = {January 2001},
publisher = {AI Access Foundation},
address = {El Segundo, CA, USA},
issn = {1076-9757},
journal = {J. Artif. Int. Res.},
}

@inproceedings{silver2021planning,
  title={Planning with learned object importance in large problem instances using graph neural networks},
  author={Silver, Tom and Chitnis, Rohan and Curtis, Aidan and Tenenbaum, Joshua B and Lozano-Perez, Tomas and Kaelbling, Leslie Pack},
  booktitle={Proceedings of the AAAI conference on artificial intelligence},
  year={2021}
}

@inproceedings{procthor,
  author={Matt Deitke and Eli VanderBilt and Alvaro Herrasti and
          Luca Weihs and Jordi Salvador and Kiana Ehsani and
          Winson Han and Eric Kolve and Ali Farhadi and
          Aniruddha Kembhavi and Roozbeh Mottaghi},
  title={{ProcTHOR: Large-Scale Embodied AI Using Procedural Generation}},
  booktitle={NeurIPS},
  year={2022},
}

@inproceedings{shridhar2020alfred,
  title={{ALFRED}: A benchmark for interpreting grounded instructions for everyday tasks},
  author={Shridhar, Mohit and Thomason, Jesse and Gordon, Daniel and Bisk, Yonatan and Han, Winson and Mottaghi, Roozbeh and Zettlemoyer, Luke and Fox, Dieter},
  booktitle={IEEE/CVF conference on computer vision and pattern recognition},
  year={2020}
}

@inproceedings{
sikes2024reducing,
title={Reducing Human-Robot Goal State Divergence with Environment Design},
author={Kelsey Sikes and Sarah Keren and Sarath Sreedharan},
booktitle={ICAPS 2024 Workshop on Human-Aware Explainable Planning},
year={2024}
}

@inproceedings{task-anticipation,
  title = {Anticipate \& {Act}: Integrating {LLM}s and Classical Planning for Efficient Task Execution in Household Environments},
  author = {Arora, Raghav and Singh, Shivam and Swaminathan, Karthik and Datta, Ahana and Banerjee, Snehasis and Bhowmick, Brojeshwar and Jatavallabhula, {Krishna Murthy} and Sridharan, Mohan and Krishna, Madhava},
  year = {2024},
  booktitle = {International Conference on Robotics and Automation (ICRA)},
}

@inproceedings{dhakal2023anticipatory,
  title =      {Anticipatory Planning: Improving Long-Lived Planning 
                by Estimating Expected Cost of Future Tasks},
  author =     {Dhakal, Roshan and Talukder, Md Ridwan Hossain 
                and Stein, Gregory J.},
  booktitle =  {International Conference on Robotics and Automation (ICRA)},
  year =       {2023},
}

@inproceedings{koppula2016anticipatory,
  title={Anticipatory planning for human-robot teams},
  author={Koppula, Hema S and Jain, Ashesh and Saxena, Ashutosh},
  booktitle={Experimental Robotics: The 14th International Symposium on Experimental Robotics},
  year={2016},
}

@inproceedings{
patel2023predicting,
title={Predicting Routine Object Usage for Proactive Robot Assistance},
author={Maithili Patel and Aswin Gururaj Prakash and Sonia Chernova},
booktitle={7th Annual Conference on Robot Learning},
year={2023}
}

@inproceedings{klassen2022planning,
  title={Planning to avoid side effects},
  author={Klassen, Toryn Q and McIlraith, Sheila A and Muise, Christian and Xu, Jarvis},
  booktitle={Proceedings of the AAAI Conference on Artificial Intelligence},
  year={2022}
}

@inproceedings{
patel2022proactive,
title={Proactive Robot Assistance via Spatio-Temporal Object Modeling},
author={Maithili Patel and Sonia Chernova},
booktitle={6th Annual Conference on Robot Learning},
year={2022}
}

@inproceedings{lowe2017maddpg,
  title={Multi-Agent Actor-Critic for Mixed Cooperative-Competitive Environments},
  author={Lowe, Ryan and Wu, Yi and Tamar, Aviv and Harb, Jean and Abbeel, Pieter and Mordatch, Igor},
  booktitle={Advances in Neural Information Processing Systems (NeurIPS)},
  year={2017}
}

@inproceedings{rashid2018qmix,
  title={QMIX: Monotonic Value Function Factorisation for Deep Multi-Agent Reinforcement Learning},
  author={Rashid, Tabish and Samvelyan, Mikayel and Schroeder, Christian and Farquhar, Gregory and Foerster, Jakob and Whiteson, Shimon},
  booktitle={International Conference on Machine Learning (ICML)},
  year={2018}
}

@article{burns2012anticipatory,
  title={Anticipatory On-Line Planning},
  journal={Proceedings of the International Conference on Automated Planning and Scheduling},
  author={Burns, Ethan and Benton, J. and Ruml, Wheeler and Yoon, Sungwook and Do, Minh},
  year={2012},
}
